%% file: manuscript.tex
\documentclass[journal,twoside]{IEEEtran}

\input{packages}
\usepackage{amsmath,amsfonts}
\usepackage{algorithmic}
\usepackage{algorithm}
\usepackage{array}
\usepackage{textcomp}
\usepackage{stfloats}
\usepackage{url}
\usepackage{verbatim}
\usepackage{graphicx}
\usepackage{cite}
\usepackage{xcolor}
\usepackage{gensymb}
\usepackage{subcaption}
\usepackage{multirow}
\begin{document}

\markboth{IEEE Robotics and Automation Letters. Preprint version. Accepted February, 2026}{Zhang \MakeLowercase{\textit{et al.}}: Human-Exoskeleton Kinematic Calibration for Dexterous Teleoperation}

\title{Human-Exoskeleton Kinematic Calibration to Improve Hand Tracking for Dexterous Teleoperation}

\author{Haiyun Zhang$^{1,\dagger}$, Stefano Dalla Gasperina$^{1,\dagger}$,  Saad N. Yousaf$^1$, Job D. Ramirez$^1$, \\
Toshimitsu Tsuboi$^2$, Tetsuya Narita$^2$, and Ashish D. Deshpande$^{1,3}$

\thanks{Manuscript received: December 5, 2025; Revised: February 5, 2026; Accepted: February 5, 2026.}
\thanks{This paper was recommended for publication by Editor Pietro Valdastri upon evaluation of the Associate Editor and Reviewers comments.}
\thanks{This work was supported by Sony Group Corporation, Tokyo, Japan.}
\thanks{$^\dagger$These authors contributed equally to this work.}
\thanks{$^{1}$Walker Department of Mechanical Engineering, The University of Texas at Austin, Austin, TX, USA.}
\thanks{$^{2}$Sony Group Corporation, Tokyo, Japan.}
\thanks{$^{3}$Meta Reality Labs Research, Redmond, WA, USA.}

\thanks{Digital Object Identifier (DOI): see top of this page.}
}



\maketitle

\begin{abstract}
Hand exoskeletons are critical tools for dexterous teleoperation and immersive manipulation interfaces, but achieving accurate hand tracking remains a challenge due to user-specific anatomical variability and donning inconsistencies. These issues lead to kinematic misalignments that degrade tracking performance and limit applicability in precision tasks. We propose a subject-specific calibration framework for exoskeleton-based hand tracking that estimates virtual link parameters through residual-weighted optimization. A data-driven approach is introduced to empirically tune cost function weights using motion capture ground truth, enabling accurate and consistent calibration across users. Implemented on the \textsc{Maestro} hand exoskeleton with seven healthy participants, the method achieved substantial reductions in joint and fingertip tracking errors across diverse hand geometries. Qualitative visualizations using a Unity-based virtual hand further demonstrate improved motion fidelity. While demonstrated on the \textsc{Maestro} exoskeleton, the framework may be extended in principle to other hand exoskeleton systems that provide comparable kinematic constraints and sensing capabilities.

\end{abstract}


\input{01_introduction}

\input{02_methods}

\input{03_experiments}

\input{04_results}

\input{05_discussion}

\bibliographystyle{ieeetr}
\bibliography{ref,ref_updated}
\end{document}

%% file: packages.tex
\usepackage{graphics} 
\usepackage{epsfig} 
\usepackage{mathptmx} 
\usepackage{times} 
\usepackage{amsmath} 
\usepackage{amssymb}  
\usepackage{verbatim} 
\usepackage{textcomp}
\usepackage{url}
\usepackage{booktabs,chemformula} 

\usepackage{gensymb}
\usepackage{multirow}

\usepackage{hyperref}

\usepackage{caption}
\usepackage{subcaption}
\usepackage{siunitx}
\usepackage{tikz}
\usepackage{wasysym}
\usepackage{svg}
\usepackage{placeins}

\usepackage{ulem}
\usepackage{soul}
\usepackage{xcolor}

\usepackage{pifont}

\usepackage[font=small]{caption}
\usepackage[outline]{contour}

\usepackage{lipsum}
\usepackage{hyperref}

\usepackage{tikz}

\usepackage[]{svg}

\usepackage{multirow} 
\usepackage{makecell} 

\usepackage{array}

\usepackage{colortbl} 

\usepackage{hhline}

\usepackage{diagbox}

\usepackage{adjustbox} 

\usepackage{cite}
\usepackage{comment}

\usepackage{siunitx}

%% file: 01_introduction.tex
\section{Introduction}
\label{sec:introduction}

Achieving human-level dexterity in robotic manipulation remains a key challenge~\cite{zhang2025dexterous}, as current systems still fall short in achieving the versatility needed for tool use and fine manipulation~\cite{yu2022dexterous,chen2023visual}.

Dexterous teleoperation, where humans control robotic hands to perform fine tasks, offers a practical pathway toward this goal, supporting applications from surgery and space exploration to robot learning from human demonstration~\cite{si2021review}.
Its success depends critically on accurate, low-latency hand tracking that reliably captures user motion.

Existing approaches fall into two main categories: vision-based systems and wearable devices~\cite{mizera2019evaluation,fu2022teleoperation}. Vision-based methods such as optical motion capture or camera-based tracking~\cite{vakunov2020mediapipe,mueller2017real}, provide non-invasive sensing but suffer from occlusions, lighting sensitivity, and workspace limits, and they lack haptic feedback for force-based interactions~\cite{moon2023hand}.

\begin{figure}[t!]
    \centering
    \begin{subfigure}[t]{0.62\columnwidth}
        \centering        
        \includegraphics[width=\textwidth]{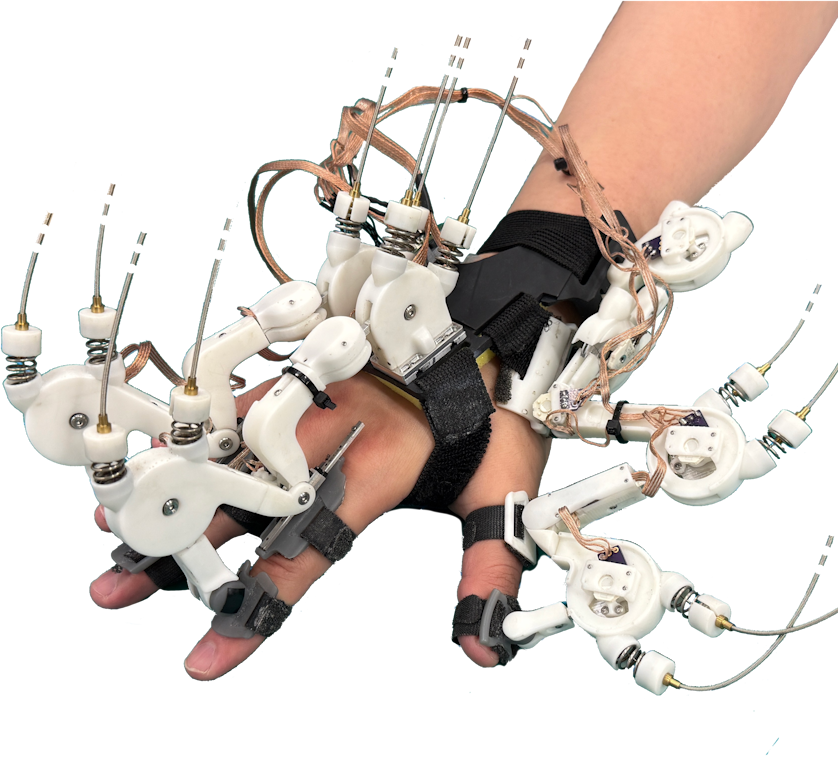}
    \end{subfigure}%
        \begin{subfigure}[t]{0.4\columnwidth}
        \centering        
        \includegraphics[width=\textwidth]{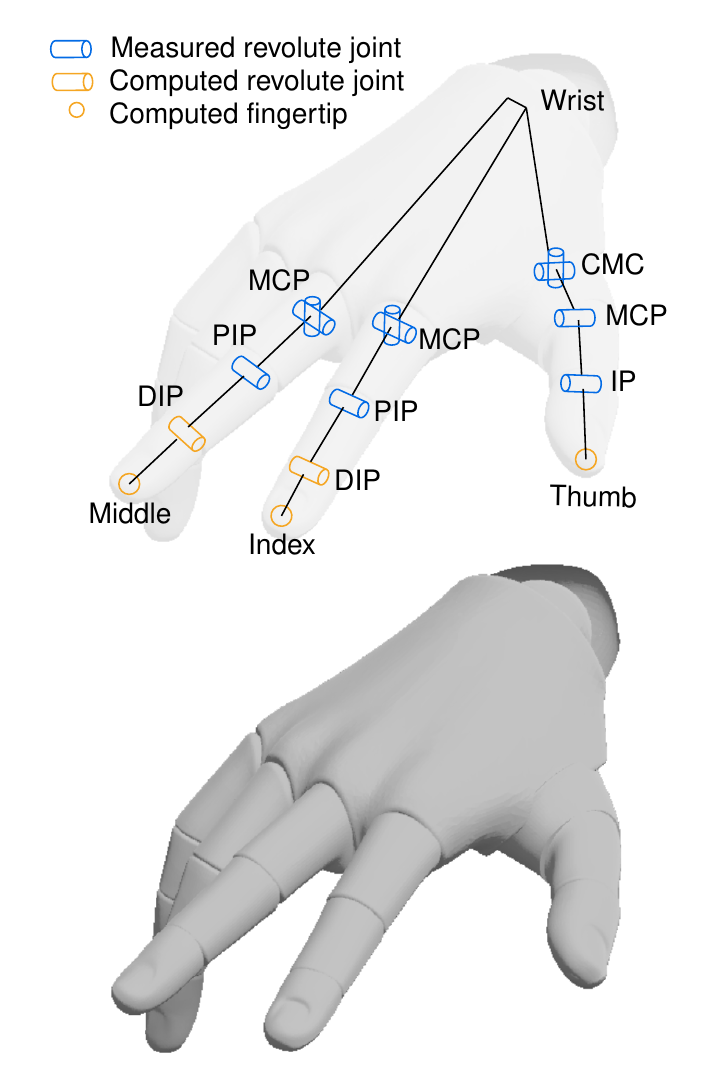}
    \end{subfigure}%
    \caption{(Left) The \textsc{Maestro} hand exoskeleton worn on the thumb, index, and middle fingers. (Right) Simplified kinematic model of the exoskeleton and corresponding rendering of the mapped virtual hand. 
    }    \label{fig1:maestro_exoskeleton}
    \vspace{-5mm}
\end{figure}

In contrast, wearable devices, such as data gloves or exoskeletons, directly measure joint angles or fingertip motion~\cite{darvish2023teleoperation,leonardis2024hand}, enabling operation in cluttered environments and providing force feedback. However, their accuracy declines due to user-specific anatomy, inconsistent donning, and slippage~\cite{xia2022hand,yousaf2023experimental} and resulting in discomfort in mechanical designs that directly align device joints with anatomical joints \cite{schiele2009ergonomics}. Self-alignment mechanisms~\cite{wei2023adaptive,zhang2025doglove} mitigate kinematic incompatibility but often introduce non-linear correspondence between device and anatomy. To complement these mechanical solutions, recent research has focused on calibration strategies that explicitly compensate for anatomical variability, donning inconsistencies, and sensor misalignment~\cite{connolly2022improving}.

Commercial devices still rely on manual, pose-based calibration routines that are labor-intensive, user-dependent, and can take up to 90 minutes per glove, limiting practicality. Heinrich \textit{et al.}~\cite{heinrich2024comparison} highlighted the trade-off between calibration accuracy and complexity, motivating faster, more generalizable approaches. Recent work by Li \textit{et al.}~\cite{li2025fsglove} further advanced shape- and subject-specific calibration through unified kinematic–anatomical models that enhance tracking accuracy and consistency across users.

Hybrid and sensor-fusion approaches have also been explored. Gosala \textit{et al.}~\cite{gosala2021self} combined visual and magnetic sensors, reducing joint-angle error to 11.4\textdegree{} but requiring both cameras and wearables. Kim \textit{et al.}~\cite{kim2023calibration} improved fingertip accuracy via data-driven optimization, though it depended on subject-specific ground-truth data.

Building on these insights, our framework estimates subject-specific kinematic parameters to establish an individualized mapping between the hand and the exoskeleton. This mapping is derived from redundant sensing and a limited set of reference poses, without requiring external cameras, sensors or ground-truth references.

Specifically, we implement and validate this approach on the \textsc{Maestro} hand exoskeleton (Fig.~\ref{fig1:maestro_exoskeleton}), which employs self-aligning closed-loop four-bar linkages for kinematic compatibility. While this architecture enhances comfort and wearability, it remains sensitive to differences in hand size, anatomy, slippage, and donning variability~\cite{yousaf2023experimental}. To address these effects, we develop a subject-specific, data-driven calibration framework that estimates virtual link parameters and residual weightings directly from user motion data, compensating for geometric and anatomical variability without requiring rigid mechanical alignment.

The calibration is formulated as an optimization problem that minimizes tracking error through a residual-weighted cost function combining joint and fingertip discrepancies. The data-driven weighting scheme allows adaptation to subject-specific biomechanics and sensor contributions, resulting in more accurate hand-pose estimation and improved tracking fidelity. We evaluate the framework on seven participants with diverse hand geometries and validate its performance both quantitatively and qualitatively using a Unity-based virtual hand visualization. Although this work focuses on the \textsc{Maestro} exoskeleton, the framework provides a systematic approach for subject-specific calibration in sensorized hand exoskeletons, provided that similar kinematic constraints and sensing capabilities are available. 



%% file: 02_methods.tex
\section{Methods}
\label{methods}

\subsection{Hardware Platform}

The \textsc{Maestro} hand exoskeleton, shown in Fig.~\ref{fig1:maestro_exoskeleton}, is used in this study as a representative case to present and validate our subject-specific calibration framework. \textsc{Maestro} enables motion tracking and haptic feedback for the thumb, index, and middle fingers via a cable-driven mechanism~\cite{agarwal2015index,agarwal2017design}. The device comprises 16 joints instrumented with rotary potentiometers: five sensors per index and middle finger and six for the thumb, each including two redundant measurements.

\subsection{Kinematic Model}

Each digit of the \textsc{Maestro} hand exoskeleton is modeled independently using a dedicated kinematic chain, reflecting the modular mechanical design of the device. These models define the mapping from exoskeleton joint measurements to anatomical joint angles.

\begin{figure*}[t!]
  \centering
  \begin{subfigure}{0.9\linewidth}
    \centering
    \includegraphics[width=\linewidth]{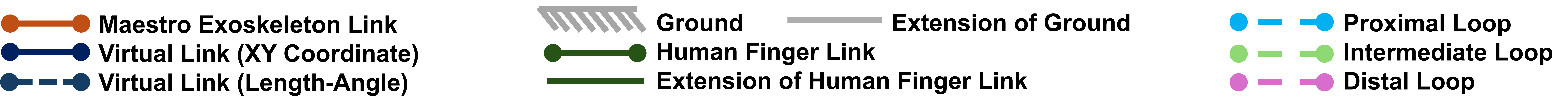}
    \caption*{} 
    \label{subfig:kin_legend}
  \end{subfigure}
  \begin{subfigure}{0.45\linewidth}
    \centering
    \includegraphics[width=\linewidth]{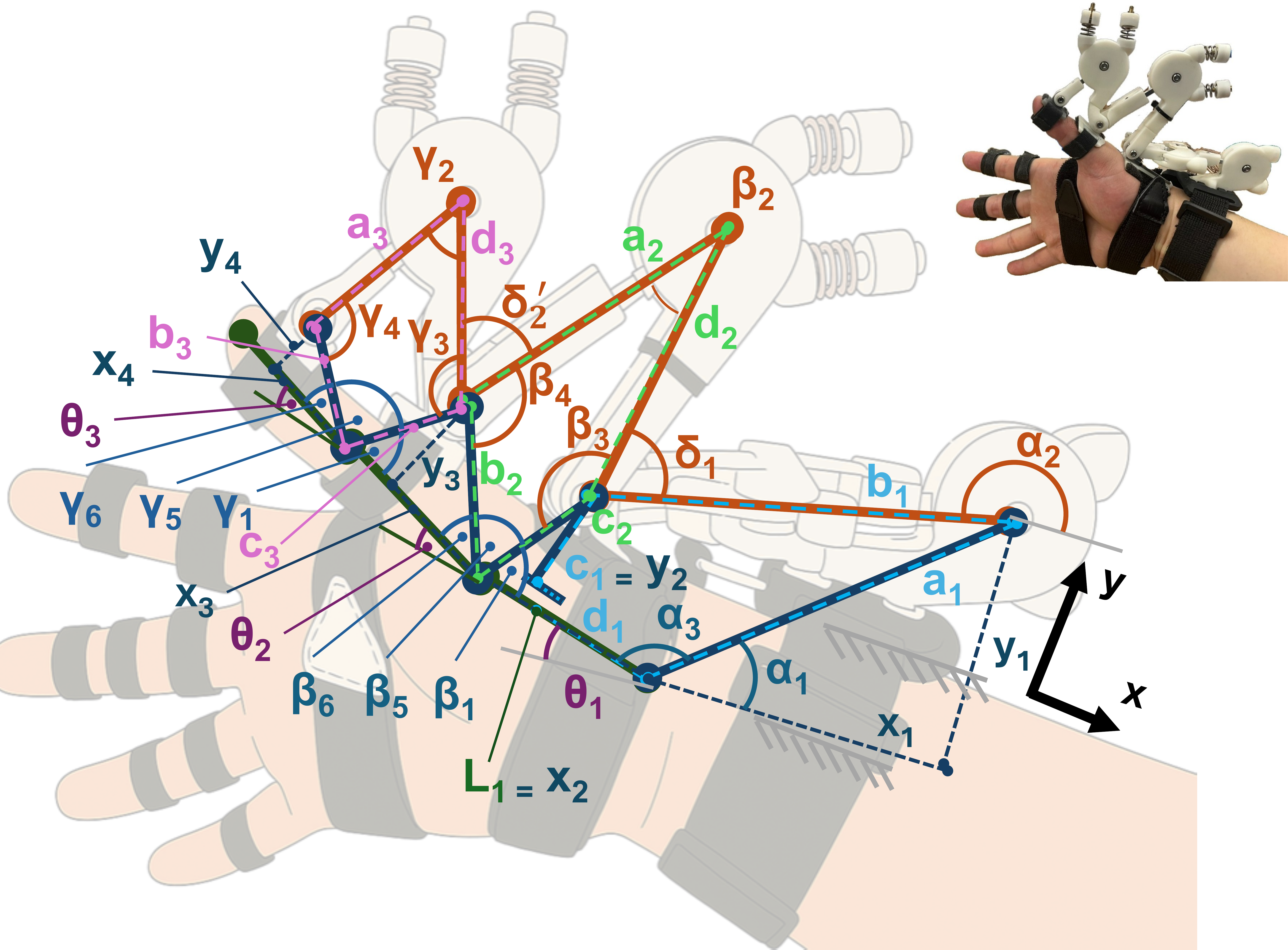}
    \caption{Thumb Kinematic Model}
    \label{subfig:kin_thumb}
  \end{subfigure}
  \hspace{0.75em}
  \begin{subfigure}{0.45\linewidth}
    \centering
    \includegraphics[width=\linewidth]{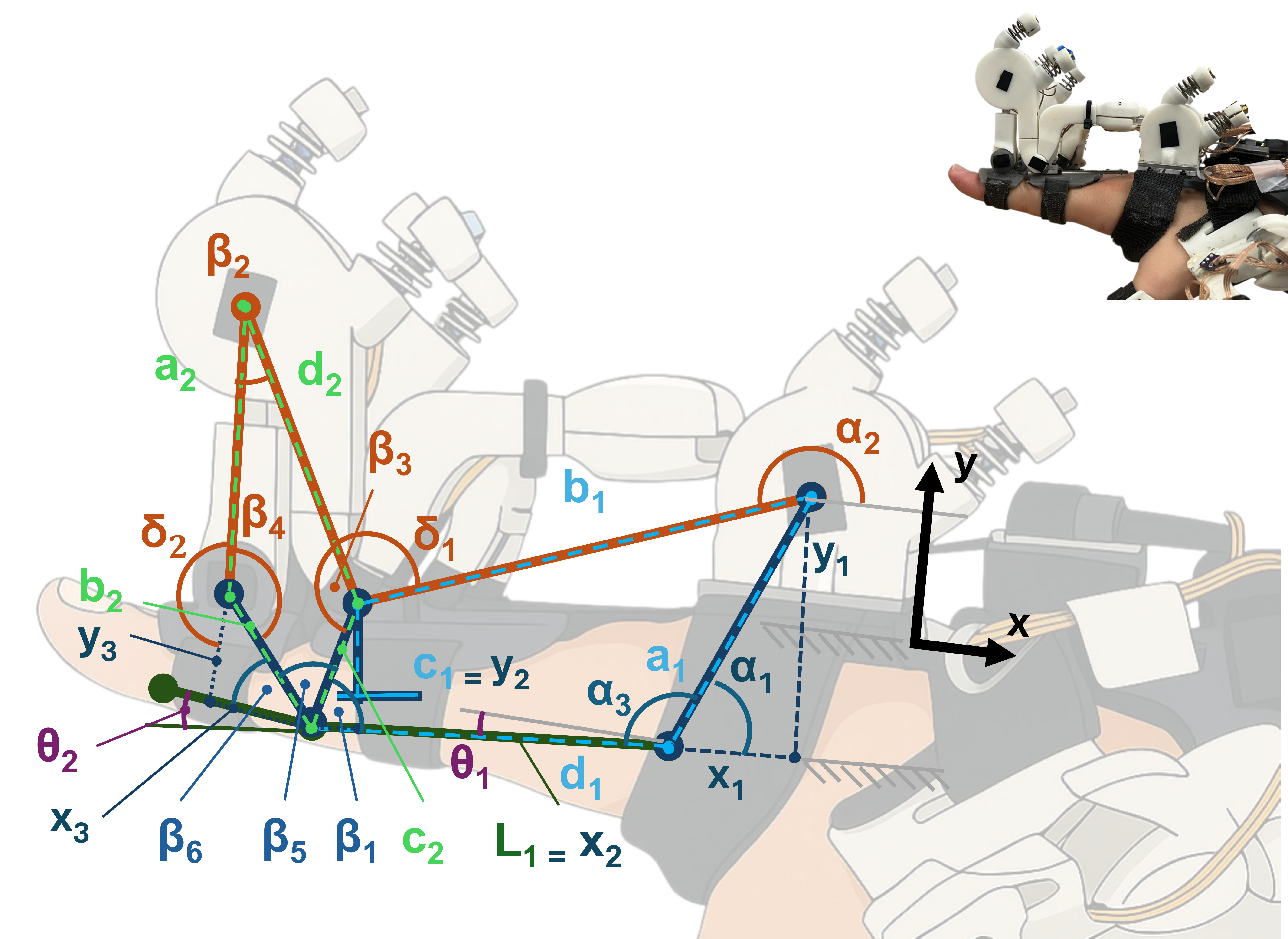}
    \caption{Index (and Middle) Finger Kinematic Model}
    \label{subfig:kin_index}
  \end{subfigure}

\caption{Kinematic model of the \textsc{Maestro} finger–exoskeleton four-bar interface. 
The thumb is equipped with redundant joints $(\delta_1,\delta_2')$ and is modeled with three kinematic loops: one RRPR loop and two RRRR loops. 
The index and middle fingers are equipped with redundant joints $(\delta_1,\delta_2)$ and are modeled with two kinematic loops: one RRPR loop and one RRRR loop. The $X$–$Y$ frame is defined such that the $X$-axis is aligned longitudinally with the base frame and the $Y$-axis is perpendicular to it. All variable definitions are provided in Tab.~\ref{tab:equations}.}
\label{fig_2:kin}
\end{figure*}

The thumb and fingers differ in mechanical structure but share several underlying kinematic equations. As shown in Fig.~\ref{fig_2:kin}, each digit consists of multiple four-bar closed-loop chains—two for the index and middle fingers, and three for the thumb. All digits follow a common layout: an initial RRPR loop, followed by one (index/middle) or two (thumb) RRRR loops. For each loop, the anatomical joint angle is estimated by modeling the combined kinematics of the exoskeleton linkages, finger segments, and virtual links.

In the following, we present the human–exoskeleton kinematic model that maps the exoskeleton’s sensorized joint angles to the corresponding anatomical joint angles. 
For the index and middle fingers, the joint angles are $(\alpha_2, \beta_2, \delta_1, \delta_2)$, where $\delta_1$ and $\delta_2$ represent the two redundant joints. Whereas for the thumb, the sensorized exoskeleton joint angles are denoted as $(\alpha_2, \beta_2, \gamma_2, \delta_1, \delta_2')$, where $\delta_1$ and $\delta_2'$ serve as redundant measurements.The corresponding anatomical joint angles are $(\theta_1, \theta_2)$ for the MCP and PIP joints of the index and middle fingers, and $(\theta_1, \theta_2, \theta_3)$ for the CMC, MCP, and IP joints of the thumb, respectively. Although similar models are defined for all digits, the thumb model is presented here for clarity. Complete loop equations and joint mappings are provided in Fig.~\ref{fig_2:kin} and Tab.~\ref{tab:equations}.

Using known geometric parameters, the input angles yield intermediate link angles through closed-form trigonometric expressions, as derived in Eqs.~\ref{eq:1st_loop}–\ref{eq:3rd_loop}. Due to the RRPR structure of the first loop, the distances $d_1$ and $c_2$ are also functions of $\alpha_2$. The analytical solutions are derived using sum-to-product trigonometric identities for compactness.

The exoskeleton includes redundant joint sensors derived from its kinematic loops, incorporated into the calibration cost to improve estimation accuracy. Each digit has two redundant sensors: $(\delta_1,\delta_2')$ for index/middle and $(\delta_1,\delta_2)$ for the thumb, as shown in Fig.~\ref{fig_2:kin}.

The first redundant joint, $\hat{\delta}_1$, is common to all digits and is defined as the angle between links $b_1$ and $d_2$, depending on both the first and second loops. 
The second redundant joint differs by digit: for the index and middle fingers, $\hat{\delta}_2$ is defined as the angle between $a_2$ and the projection of $b_2$ along the $y$-axis ($y_3$), influenced by the second loop geometry $f(\beta)$; for the thumb, $\hat{\delta}_2'$ is defined as the angle between $a_2$ and $d_3$, depending on the second and third loops $f(\beta,\gamma)$. 
Thus, $\delta_2$ denotes the second redundant joint of the index/middle fingers, while $\delta_2'$ denotes the second redundant joint of the thumb.



These estimated redundant angles are computed using geometric constraints, as:

\begin{equation}
\hat{\delta}_1 =2\pi + \theta_1 + \beta_1 - \alpha_2 - \beta_3
\end{equation}
\begin{equation}
\hat{\delta}_2 = 2\pi - \beta_4 - (\pi/2 - \beta_5)
\end{equation}
\begin{equation}
\begin{split}
    \hat{\delta}_2'= 2\pi - \beta_4 - (\pi/2 - \beta_6) - \gamma_3 - (\pi/2 - \gamma_1)
\end{split}
\end{equation}
where the intermediate variables are presented in Tab.~\ref{tab:equations}.
These angles, estimated from loop geometry, provide additional constraints used in the calibration cost function.

While the kinematic model defines a consistent mapping from exoskeleton to anatomical joints, this mapping is sensitive to inter-subject variability such as hand anatomy, size, and donning configuration. These factors influence the virtual link geometry and can introduce tracking inaccuracies if left uncalibrated. The following section analyzes how variations in key kinematic parameters affect tracking performance, motivating the need for a subject-specific calibration framework.

\input{table1_equations}

\subsection{Kinematic Parameters Sensitivity Analysis}
\label{sec:sensitivity_analysis}

To motivate the need for a calibration procedure that adjusts the kinematic parameter values, we first analyzed the sensitivity of fingertip position to perturbations in the virtual link parameters. The goal was to demonstrate that even small errors in these parameters can lead to significant inaccuracies in fingertip tracking.
Using the kinematic model of the index finger, each of the six 2D virtual link coordinates $\{x_1, y_1, x_2, y_2, x_3, y_3\}$ was perturbed independently within $\pm$ 10\% of its nominal value in simulation, and the resulting deviation in fingertip position was computed. 

\begin{figure}[t!]
  \centering
  \subfloat[Parameter $x_1$] {
    \includegraphics[width=0.3\linewidth]{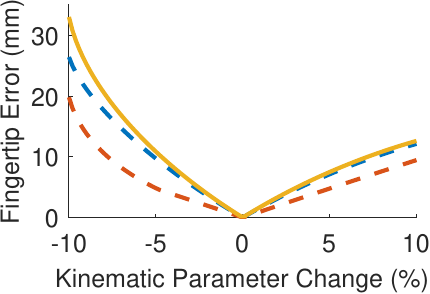}
    \label{subfig:x1}}
  \subfloat[Parameter $x_2$] {
    \includegraphics[width=0.3\linewidth]{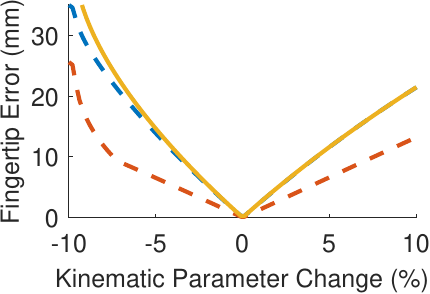}
    \label{subfig:x2}}
  \subfloat[Parameter $x_3$] {
    \includegraphics[width=0.3\linewidth]{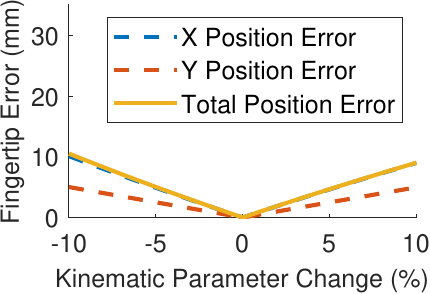}
    \label{subfig:x3}}
  \hfill
  \subfloat[Parameter $y_1$] {
    \includegraphics[width=0.3\linewidth]{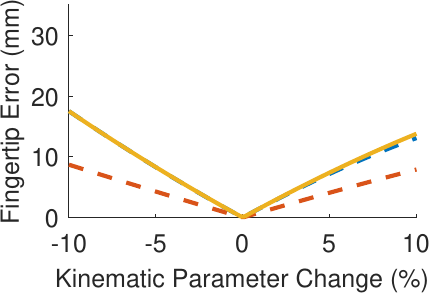}
    \label{subfig:y1}}
  \subfloat[Parameter $y_2$] {
    \includegraphics[width=0.3\linewidth]{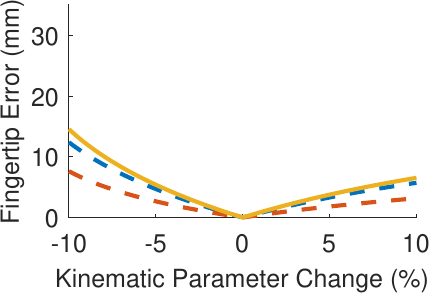}
    \label{subfig:y2}}
  \subfloat[Parameter $y_3$] {
    \includegraphics[width=0.3\linewidth]{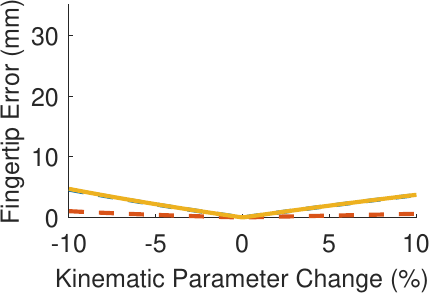}
    \label{subfig:y3}}
  \caption{Sensitivity analysis in simulation of fingertip position error (mm) as a function of perturbations in select parameters (\% variation).}
  \label{fig:sensitivity_analysis_results}
  \vspace{-3mm}
\end{figure}

The results, shown in Fig.~\ref{fig:sensitivity_analysis_results}, indicate that proximal parameters, particularly $x_1$ and $y_1$, exert the strongest influence on fingertip position, with 10\% perturbations causing deviations of up to 30\,\si{mm}. 
In contrast, distal parameters such as $x_3$ and $y_3$ had only minor effects, typically below 3–5\,\si{mm}. 

Fingertip tracking was more sensitive to horizontal ($x$) perturbations, consistent with common horizontal sliding of the device on the back of the hand during use.
This finding underscores the need for calibration that optimizes virtual link parameters to maintain fingertip tracking accuracy.

\subsection{Human-Exoskeleton Calibration Framework}
\label{sec:calibration}

Accurate hand tracking in exoskeletons like \textsc{Maestro} is challenging due to non-anatomical joint mappings and user-specific virtual links. These links vary with hand size and donning and cannot be directly measured. We address this by calibrating them from motion data to improve tracking.

\begin{figure}[t!]
    \centering
    \includegraphics[width=\linewidth]{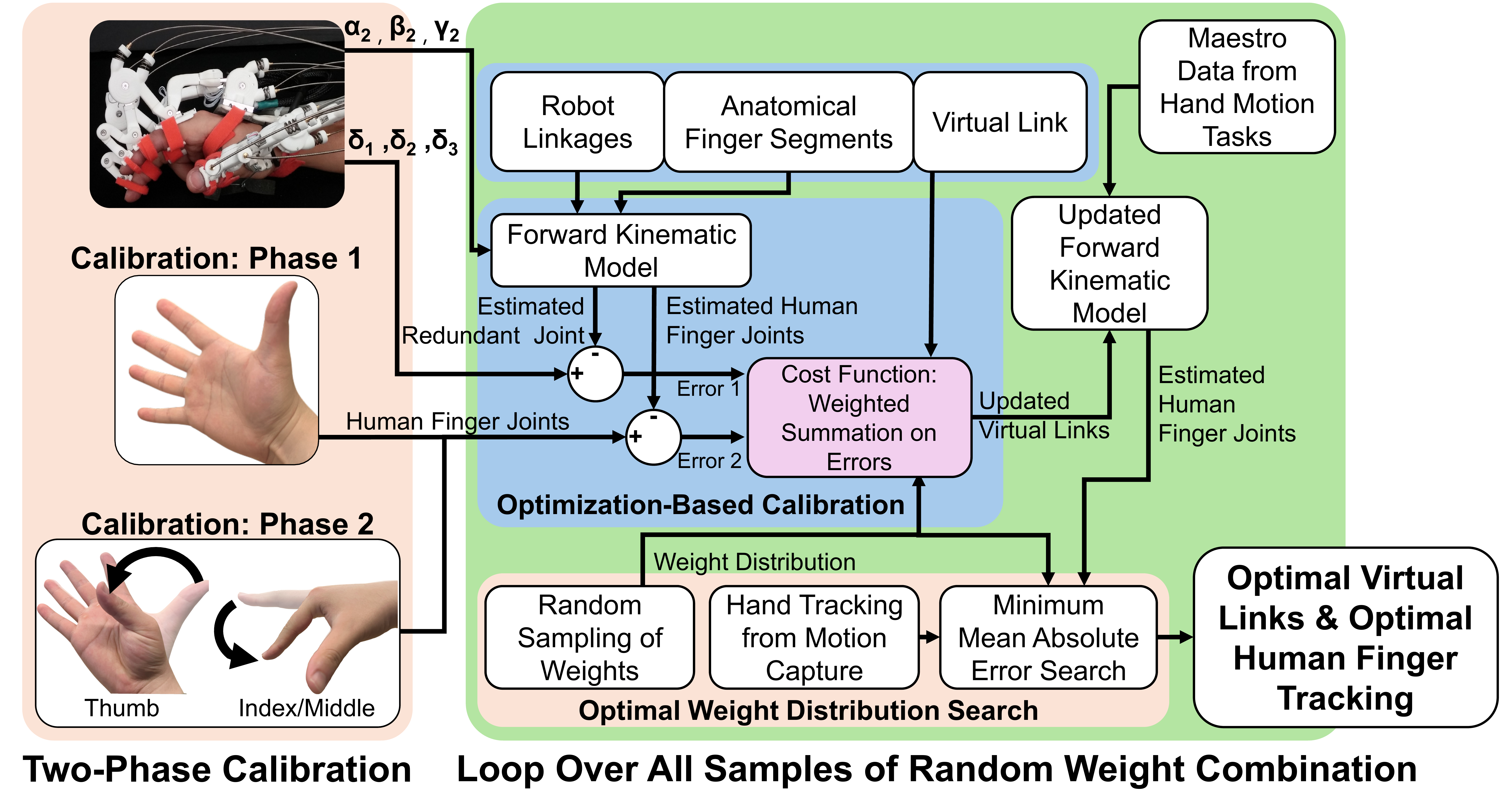}
    \caption{Visualization of the kinematic parameter calibration with weighted optimization. The human user performs a two-phase calibration, and the weight distribution is adjusted based on human data.}
    \label{fig_3:calibration}
    \vspace{-5mm}
\end{figure}

The goal of the calibration process is to optimize subject- and donning-specific virtual link parameters (e.g., a$_1$, c$_1$, b$_2$, c$_3$) by minimizing discrepancies between estimated ($\hat{\theta}_n$, $\hat{\delta}_n$) and measured joint angles ($\delta_{n,\text{ref}}$) or known configurations ($\theta_{n,\text{ref}}$). The inputs to the calibration are the measured joint angles, while the outputs are the estimated virtual links, which generate updated estimates of the user’s joint angles. The full procedure is illustrated in Fig.~\ref{fig_3:calibration}.

In phase one ($\phi_1$), subjects fully extend their fingers to form a flat-hand posture and hold it for a few seconds while static joint measurements are recorded. This configuration imposes known anatomical constraints, which serve as reference angles ($\theta_\text{ref}$ and $\delta_\text{ref}$) for calibration: all finger joints are assumed to be at $0\degree$, except for the thumb CMC flexion/extension, which we fixed at 70\textdegree when participants opened their hands in a flat posture. Although this value is not anatomically identical across individuals, it provides a consistent reference across subjects and ensures the calibration problem remains well-posed.

In phase two ($\phi_2$), subjects perform isolated flexion of the MCP joints of the thumb, index, and middle fingers while keeping the IP (thumb) and PIP (index/middle) joints extended. 
Here, the MCP reference angles are unknown and serve as optimization targets, while the IP/PIP joints are again assumed to remain at $0\degree$. 


The two-phase protocol enables global alignment through the flat-hand posture and enforces dynamic consistency via isolated MCP flexion. Together, these conditions provide complementary constraints that help the calibration generalize across configurations. An optimization routine then updates the virtual link parameters by minimizing the discrepancy between measured redundant joint angles (or reference postures) and those predicted by the kinematic model. 


Virtual link parameters can be expressed in two equivalent forms: a length–angle form $(a_1,\alpha_1,c_2,\beta_1,b_2,\beta_6,b_3,\gamma_6)$ for kinematic modeling and an XY-coordinate form $(x_1,y_1,x_2,y_2,x_3,y_3,x_4,y_4)$ for optimization. The two are related through trigonometric identities; the former enables closed-form angle computation, while the latter ensures dimensional consistency during optimization.

The calibration minimizes a weighted sum of squared errors between model-predicted and reference joint angles, where each error term $\Delta$ represents the discrepancy between estimated and measured values for anatomical or redundant joints, respectively (Eqs.~\ref{eq:cost_function_1}-\ref{eq:cost_function_2}).

\begin{equation}
\label{eq:cost_function_1}
\Delta\theta_{n,\phi_n}^{(i)} = \hat{\theta}_n^{(i)} - \theta_{n,\text{ref}}
\end{equation}
for anatomical joints with known reference, where the superscript $i$ identifies samples over the calibration period, $n$ is the joint index, $\phi_n$ is the calibration phase, and $\theta_{\text{ref}}$ is the measured or known reference angle (e.g., 0° during hand flattening)
\begin{equation}
\label{eq:cost_function_2}
\Delta\delta{n,\phi_n}^{(i)} = \hat{\delta}_n^{(i)} - \delta_{n,\text{ref}}^{(i)}
\end{equation}
for redundant joints with angular sensors, where ${\delta}_{\text{ref}}$ is the reference angle measured by the redundant sensors.


For the index and middle fingers, the cost function is:

\begin{equation}
\small
\begin{aligned}
f_{1}(\mathbf{p}) = \frac{1}{N} \sum_{i=1}^{N} \Big(&
w_1(\Delta\theta_{1,\phi_1}^{(i)})^2 +
w_2(\Delta\theta_{2,\phi_1}^{(i)})^2 +
w_3(\Delta\delta_{1,\phi_1}^{(i)})^2 + \\
& w_4(\Delta\delta_{2,\phi_1}^{(i)})^2 +
w_5(\Delta\theta_{2,\phi_2}^{(i)})^2 +
w_6(\Delta\delta_{1,\phi_2}^{(i)})^2 \Big)
\end{aligned}
\label{eq:cost}
\end{equation}
where $\mathbf{p} = (x_1, y_1, x_2, y_2, x_3, y_3)$, $w_k$ are weights and $N$ is the number of samples. To ensure a well-posed optimization, the number of error terms must match the number of virtual link parameters. This condition ensures that each parameter can be uniquely estimated from independent error constraints without introducing underdetermined or redundant solutions.

To ensure a well-posed optimization, the number of error terms must match the number of virtual link parameters. For example, six parameters require six independent error terms. 
Although both $\Delta\delta_{1,\phi_2}$ and $\Delta\delta_{2,\phi_2}$ are theoretically valid, $\Delta\delta_{1,\phi_2}$ was used since $\delta_{2,\phi_2}$ was prone to interface tilting and inconsistent contact during flexion.


Similarly, the cost function of the thumb is defined as:
\begin{equation}
\small
\begin{aligned}
f_2(\mathbf{q}) = \frac{1}{N} \sum_{i=1}^{N} \Big( 
w_1(\Delta\theta_{1,\phi_1}^{(i)})^2 +
w_2(\Delta\theta_{2,\phi_1}^{(i)})^2 +
w_3(\Delta\theta_{3,\phi_1}^{(i)})^2 + \\
w_4(\Delta \delta_{1,\phi_1}^{(i)})^2 + 
w_5(\Delta \delta_{2,\phi_1}'^{(i)})^2 + 
w_6(\Delta\theta_{3,\phi_2}^{(i)})^2 + \\[1ex]
w_7(\Delta \delta_{1,\phi_2}^{(i)})^2 + 
w_8(\Delta \delta_{2,\phi_2}'^{(i)})^2 \Big)
\end{aligned}
\label{eq:cost}
\end{equation}
where $\mathbf{q} = (x_1, y_1, x_2, y_2, x_3, y_3, x_4, y_4)$. Subsequently, the global optima of the two problems are obtained as:
\begin{equation}
\mathbf{p}^* = \arg\min_{\mathbf{p}} f_1(\mathbf{p}), \quad
\mathbf{q}^* = \arg\min_{\mathbf{q}} f_2(\mathbf{q})
\end{equation}

The kinematic calibration procedure relies on optimizing a set of virtual link parameters by minimizing a weighted sum of joint angle errors with the solver $fminunc$ from MATLAB. The performance of this optimization depends not only on the quality of the input data but also on how the error terms are weighted in the cost function. While the simulation-based sensitivity analysis, presented in~\ref{sec:sensitivity_analysis}, offers a first-principles rationale for prioritizing proximal parameters, it does not capture real-world factors such as model inaccuracies and inter-subject variability. To address this, we performed a data-driven weight refinement using motion-capture validation. 

Although illustrated on the \textsc{Maestro} four-bar design, the calibration principle can be extended in principle to other hand–exoskeleton architectures that provide sufficient independent constraints from redundant sensing, loop closures, or reference poses. In practice, each device still requires its own cost-function formulation and parameter tuning. Further discussion on scope and limitations is provided in Section~\ref{sec:discussion}.

%% file: table1_equations.tex
\begin{table*}[h!]
\small
\centering
\vspace{1em}
\begin{tabular}{@{}p{0.33\textwidth}@{} p{0.33\textwidth}@{} p{0.33\textwidth}@{}}

\toprule
\textbf{Proximal Loop (thumb and index)} & \textbf{Intermediate Loop (thumb and index)} & \textbf{Distal Loop (thumb only)} \\

\midrule

\textbf{Input:} $\alpha_2$ &
\textbf{Input:} $\beta_2,\delta_1, \delta_2$ & 
\textbf{Input:} $\gamma_2, \delta_2'$ \\

\textbf{Parameters:} a$_1$,b$_1$,c$_1,L_1,\alpha_1$ &
\textbf{Parameters:} $a_2,b_2,d_2,\beta_6$ & \textbf{Parameters:} $a_3,b_3,c_3,d_3,\gamma_1,\gamma_6$ \\

\textbf{Variables:} d$_1,\alpha_3$ &
\textbf{Variables:} c$_2,\beta_1,\beta_3,\beta_4,\beta_5$ &
\textbf{Variables:} $\gamma_3$,$\gamma_5$\\

\textbf{Output:} $\theta_1$ & 
\textbf{Output:} $\theta_2,\hat{\delta}_1,\hat{\delta}_2$ (index/middle only) & 
\textbf{Output:} $\theta_3,\hat{\delta}_2' $ (thumb only) \\ 

\midrule
\textbf{Solution:} &
\textbf{Solution:} &
\textbf{Solution:} \\

\begin{minipage}[t]{0.25\textwidth}
\vspace{-1.5ex}
\footnotesize
\begin{equation}
\theta_1 = -\pi + \alpha_3
\label{eq:1st_loop}
\end{equation}
\end{minipage} &

\begin{minipage}[t]{0.25\textwidth}
\footnotesize
\vspace{-1.5ex}
\begin{equation}
\theta_2 = \beta_1 + \beta_5 + \beta_6 - \pi
\end{equation}
\end{minipage} &

\begin{minipage}[t]{0.25\textwidth}
\footnotesize
\vspace{-1.5ex}
\begin{equation}
\theta_3 = \gamma_1 + \gamma_5 + \gamma_6 - \pi
\label{eq:3rd_loop}
\end{equation}
\end{minipage} \\[1.5ex]





\midrule
\textbf{Where:} & 
\textbf{Where:} & 
\textbf{Where:} \\

\begin{minipage}[t]{0.3\textwidth}
\footnotesize
$\begin{array}{rl}

d_1 &= \sqrt{\scriptstyle a_1^2 + b_1^2 - c_1^2 + 2a_1b_1 \cos(\alpha_2 - \alpha_1)} \\[1ex]

\alpha_3 &= -2\arctan 
\left( 
\tiny{
\frac{
\begin{array}{@{}c@{}}
b_1 - 2d_1 + 2a_1 \cos(\alpha_2) \cos(\alpha_1) \\
+ 2(b_1 - d_1) \cos(\alpha_2) + 2a_1 \cos(\alpha_1) \\
+ b_1 \cos(2\alpha_2)
\end{array}
}{
\begin{array}{@{}c@{}}
(2 \cos(\alpha_2) + 1) \cdot \\
\left( -c_1 + a_1 \sin(\alpha_1) + b_1 \sin(\alpha_2) 
\scriptsize
\right)
\end{array}}}
\right) 
\end{array}$
\end{minipage} &

\begin{minipage}[t]{0.3\textwidth}
\footnotesize	$\begin{array}{rl}
c_2 &= \sqrt{\scriptstyle{c_1^2 + (L_1 - d_1)^2}} \\[1ex]

\beta_1 &= \arctan\left(\frac{c_1}{L_1 - d_1}\right)\\[1ex]

\beta_3 &= \arcsin\left(\frac{a_2 \sin(\beta_2)}{\sqrt{a_2^2 + d_2^2 - 2a_2 d_2 \cos(\beta_2)}}\right) \\[3ex]

\beta_4 &= \arccos\left(\frac{c_2^2 + d_2^2 - (a_2^2 + b_2^2) - 2c_2 d_2 \cos(\beta_2)}{-2a_2 b_2}\right) \\[3ex]

\beta_5 &= \arccos\left(\frac{a_2^2 + d_2^2 - (b_2^2 + c_2^2) - 2a_2d_2 \cos(\beta_2)}{-2b_2c_2}\right)\\[2ex]

&+ \arcsin\left(\frac{b_2 \sin(\beta_6)}{\sqrt{b_2^2 + c_2^2 - 2b_2 c_2 \cos(\beta_6)}}\right)

\end{array}$
\end{minipage} &

\begin{minipage}[t]{0.3\textwidth}
\footnotesize	$\begin{array}{rl}

\gamma_3 &= \arcsin\left(\frac{a_3\cdot \sin(\gamma_2)}{\sqrt{a_3^2+d_3^2-2a_3d_3\cos(\gamma_2)}}\right)  + \\

&+ \arcsin\left(\frac{b_3\cdot \sin(\gamma_5)}{\sqrt{b_3^2+c_3^2-2b_3c_3\cos(\gamma_5)}}\right)  \\

\gamma_5 &= \arccos\left(\frac{a_3^2 + d_3^2 - (b_3^2 + c_3^2) - 2a_3d_3 \cdot \cos(\gamma_2)}{-2b_3c_3}\right) \\

\end{array}$
\end{minipage} \\

\bottomrule
\end{tabular}
\caption{Summary of closed-loop kinematic models for computing human finger joint angles from exoskeleton sensor inputs. Each loop models one anatomical joint, using a single exoskeleton joint angle as input and incorporating exoskeleton linkages, anatomical finger segments, and virtual link lengths. The 1st and 2nd loops are used for the index, middle, and thumb digits; the 3rd loop is unique to the thumb.}
\label{tab:equations}
\end{table*}









%% file: 03_experiments.tex
\section{Human-Subject Experiments}
\label{sec:human_experiments}

\begin{figure}[t!]
  \centering
  \subfloat[Thumb markers]{
    \includegraphics[width=0.38\linewidth]{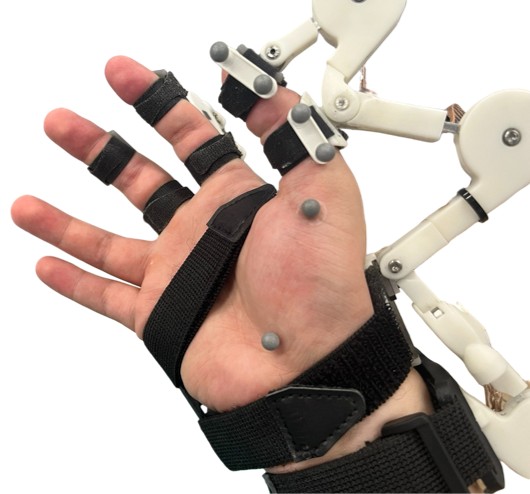}
    \label{fig_5a:mocap1}}
  \subfloat[Index markers] {
    \includegraphics[width=0.35\linewidth]{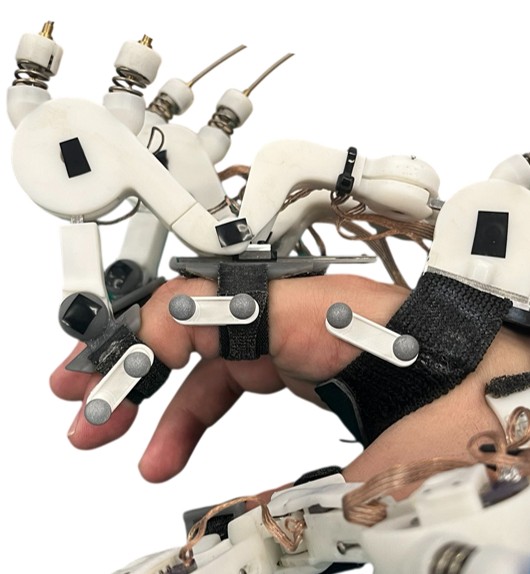}
    \label{fig_5b:mocap2}}
    \hfill
    \par\medskip
  \subfloat[Experiment workflow] {
    \includegraphics[width=0.47\textwidth]{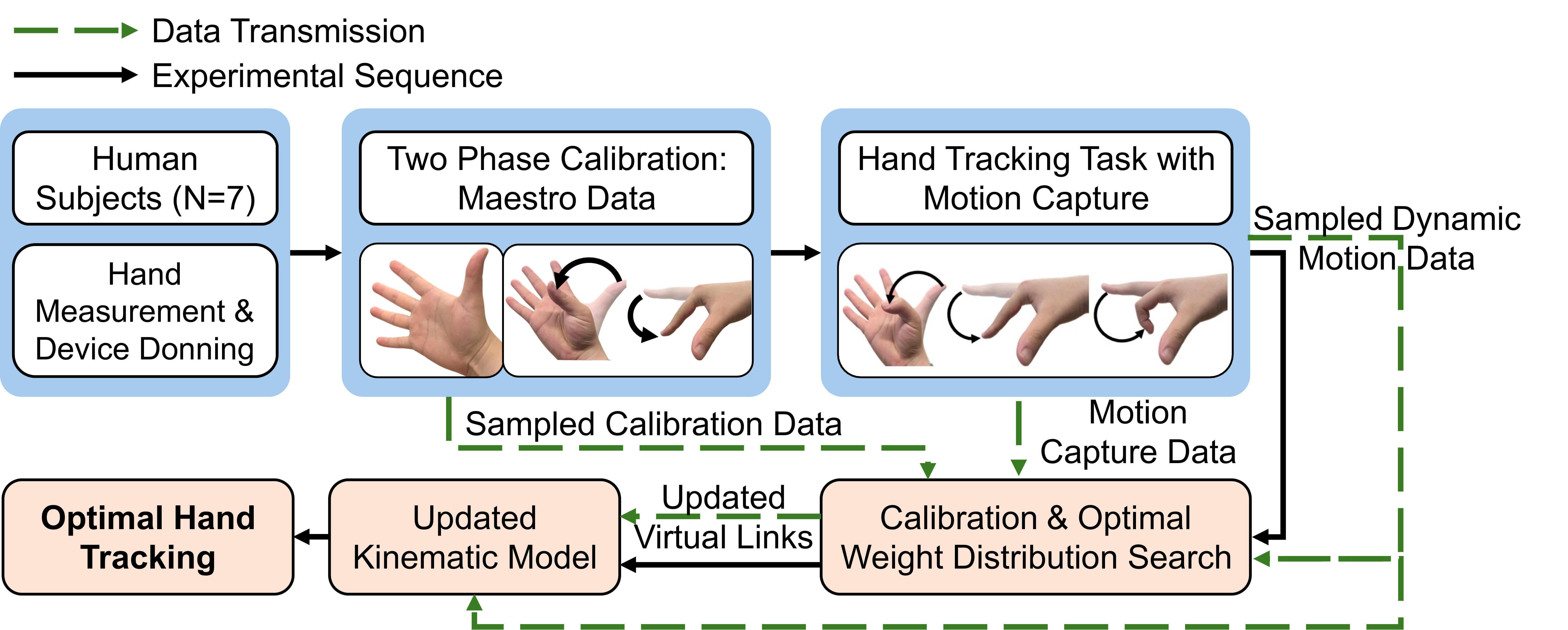}
    \label{fig_5c:expt_workflow}
    }
    \caption{Experimental setup and workflow. (a) Motion capture markers placement on the thumb, (b) on the index finger, and (c) experiment workflow for human subject testing.}
    \label{fig_4:experiment}
    \vspace{-4mm}
\end{figure}

Human-subject experiments were conducted to empirically optimize the calibration cost function weights and evaluate hand tracking performance. The study was approved by the University of Texas Institutional Review Board (IRB ID: STUDY00002527). Seven participants with varying hand sizes (Table~\ref{tab:results}) wore the exoskeleton and completed the protocol shown in Fig.~\ref{fig_5c:expt_workflow}. Reflective markers were placed on anatomical landmarks of the thumb and index finger for motion capture, as shown in Fig.~\ref{fig_5a:mocap1}-\ref{fig_5b:mocap2}.

First, participants performed the two-phase calibration routine: (i) full finger extension and (ii) MCP flexion of the thumb, index, and middle fingers with IP/PIP joints extended, as described in Sec.~\ref{sec:calibration}. These postures provided constrained joint trajectories for estimating subject-specific kinematic parameters.

Second, during the hand tracking task, participants performed three representative finger motions: (i) combined thumb MCP and IP flexion, (ii) index MCP flexion with PIP extended, and (iii) combined index MCP and PIP flexion, as shown in Fig.~\ref{fig_4:experiment}. These motions were selected to isolate individual joint estimation performance and were performed six times. \textsc{Maestro} sensor data and motion capture trajectories were recorded simultaneously and synchronized during post-processing. Due to limited visibility and occlusion when wearing the device, the analysis was limited to MCP and IP joints of the thumb, and MCP and PIP joints of the index finger. Because the index and middle fingers share an identical mechanical and kinematic design, calibration and analysis were performed on the index finger only. 

\vspace{-2ex}
\subsection{Quantitative Analysis}
\label{subsec:quantitative_analysis}

\subsubsection*{Optimal Weight Search} \label{sec:human_experiments_a}
To enhance calibration robustness against modeling uncertainties and local minima, cost function weights ($w_k$) were empirically optimized using a data-driven approach. Each subject performed dynamic finger movements while wearing the \textsc{Maestro} exoskeleton, during which a random search sampled candidate weight combinations for the cost function, and the resulting joint angle estimates were compared against ground-truth trajectories.
 For each subject, 500 candidate weight combinations ($w_k \in [0, 10]$) were randomly sampled.  Each weight distribution was used to calibrate virtual link parameters from the two-phase calibration data, and compute the corresponding joint angles and fingertip positions.
Motion capture trajectories served as ground truth to identify the weights that minimize the mean absolute error (MAE) in joint angle tracking. The weight distribution yielding the lowest error per subject was selected as optimal. Final weights were obtained by averaging these optimal weights across all participants and applied consistently in subsequent analyses.

\subsubsection*{Performance Validation} Using the averaged optimal weights, each participant's virtual link parameters were calibrated based on their two-phase calibration data. The resulting optimized kinematic model was used to estimate joint angles from \textsc{Maestro} sensor measurements during dynamic tasks. These estimates were quantitatively validated against motion capture ground truth by computing the mean absolute error (MAE) per joint, averaged across repetitions and subjects.

  \vspace{-2ex}
\subsection{Qualitative Analysis}
\label{subsec:qualitative_eval}

After verifying the quantitative performance of our exoskeleton-based hand tracking system, we qualitatively assessed how well the calibrated virtual hand resembled the user’s real hand motion in a virtual environment. A custom Unity hand model was developed using direct linear joint-to-joint mapping from the \textsc{Maestro}-estimated angles. DIP joints were inferred from corresponding PIP angles via biomechanical coupling, and all other joints were fixed. 
The visualization used anatomically plausible joint limits and averaged bone lengths across subjects and conditions, to isolate the effect of calibration. 
Participants reproduced representative static and dynamic hand gestures (e.g., pinching) under calibrated and uncalibrated conditions. These gestures were used to generate side-by-side visualization for offline inspection.

%% file: 04_results.tex
\section{Results}
\label{sec:results}

\subsection{Quantitative Results}
\subsubsection*{Optimal Weight Search}

\begin{figure}[t!]
    \centering
      \subfloat[Thumb Weights Distribution] {
        \includegraphics[width=0.44\linewidth]{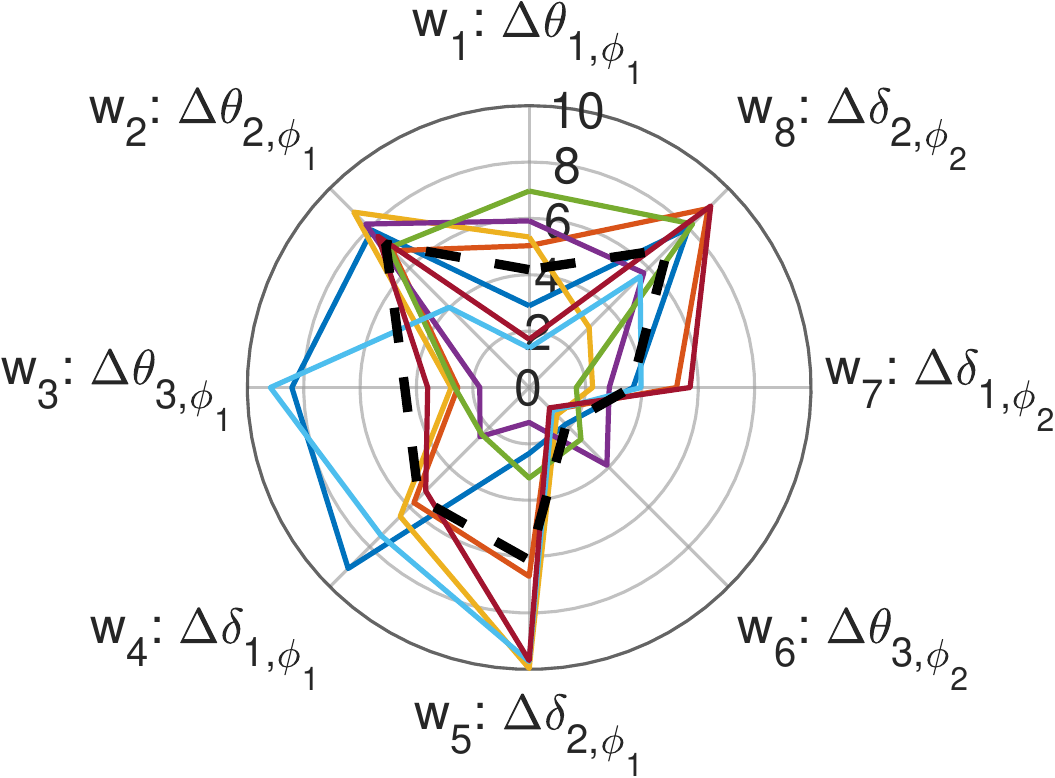}
        \label{fig:weight_thumb}}
      \subfloat[Index Weights Distribution] {
        \includegraphics[width=0.52\linewidth]{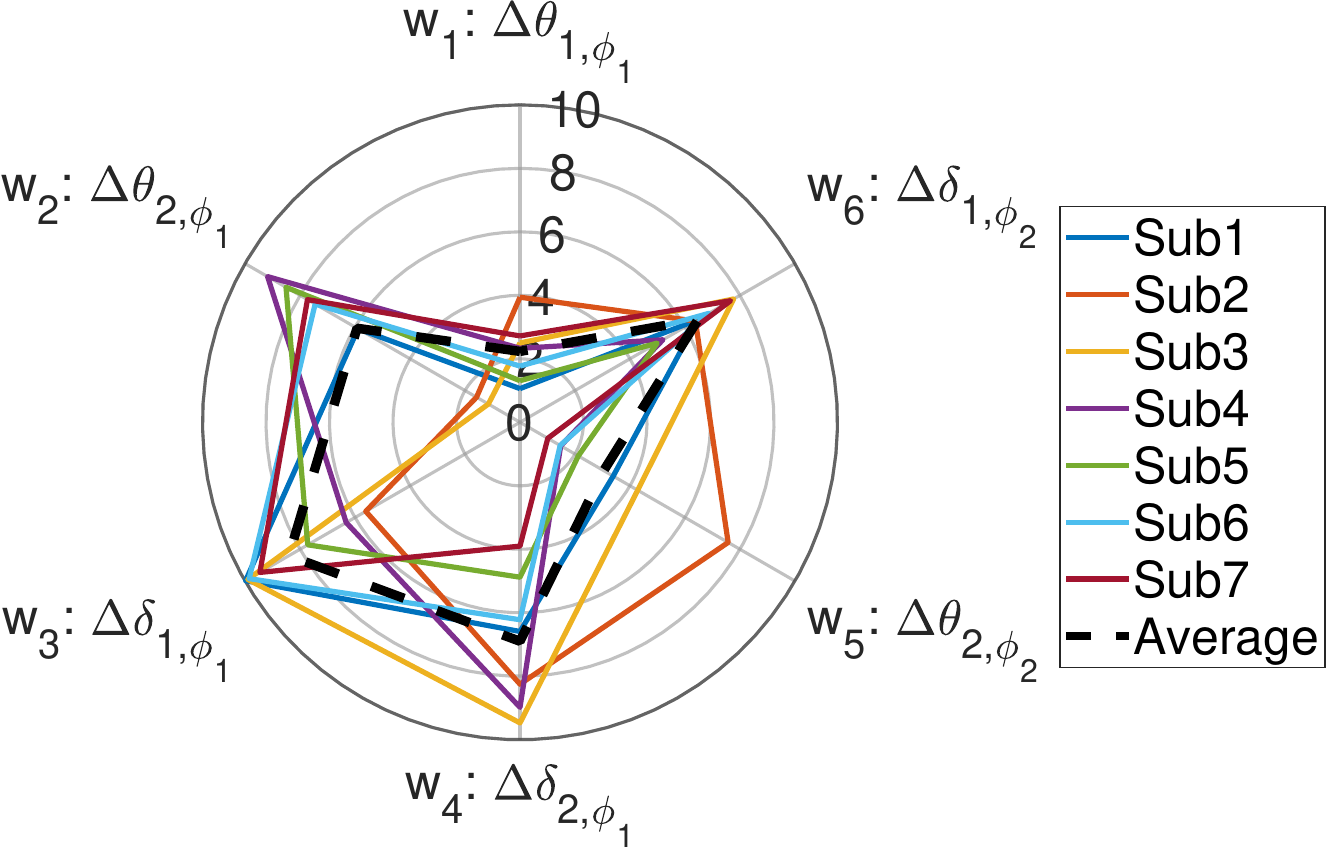}
        \label{fig:weight_index}}
    \caption{Subject-specific optimized weights distributions for thumb and index finger calibration. The dashed black line represents the averaged distribution across all subjects.}
    \label{fig:weight}
    \vspace{-1em}
\end{figure}

The cost function weights used in the calibration routine were obtained by averaging subject-specific optimal distributions, each derived by minimizing the mean absolute joint angle error relative to motion capture ground truth.
The resulting thumb and index finger distributions across all seven subjects, along with the averaged weights, are shown in Fig.~\ref{fig:weight}.
While the individual profiles exhibit broadly consistent contours, indicating robustness in the optimal weights distribution, some outliers are observed. These deviations likely stem from subject-specific differences in hand size, donning variability, or movement variability during the calibration trials. 
The averaged results for the thumb (Fig.~\ref{fig:weight_thumb}) indicate higher optimal weights for $w_2$ and $w_8$, suggesting that the calibration is more sensitive to parameters contributing to $\theta_2$ and $\delta_2$, both related to the thumb MCP joint. In contrast, the index finger results (Fig.~\ref{fig:weight_index}) show greater weights for $w_3$ and $w_4$, highlighting the sensitivity of the calibration to $\delta_1$ and $\delta_2$, associated with the PIP joints of the index and middle fingers. Overall, the optimized weights emphasize the second kinematic loop parameters---particularly the thumb MCP and finger PIP joints---as key contributors to accurate hand tracking.

\subsubsection*{Performance Validation}

\begin{figure}[t!]
    \centering
    \vspace{0.5em}
    \includegraphics[width=0.9\linewidth]{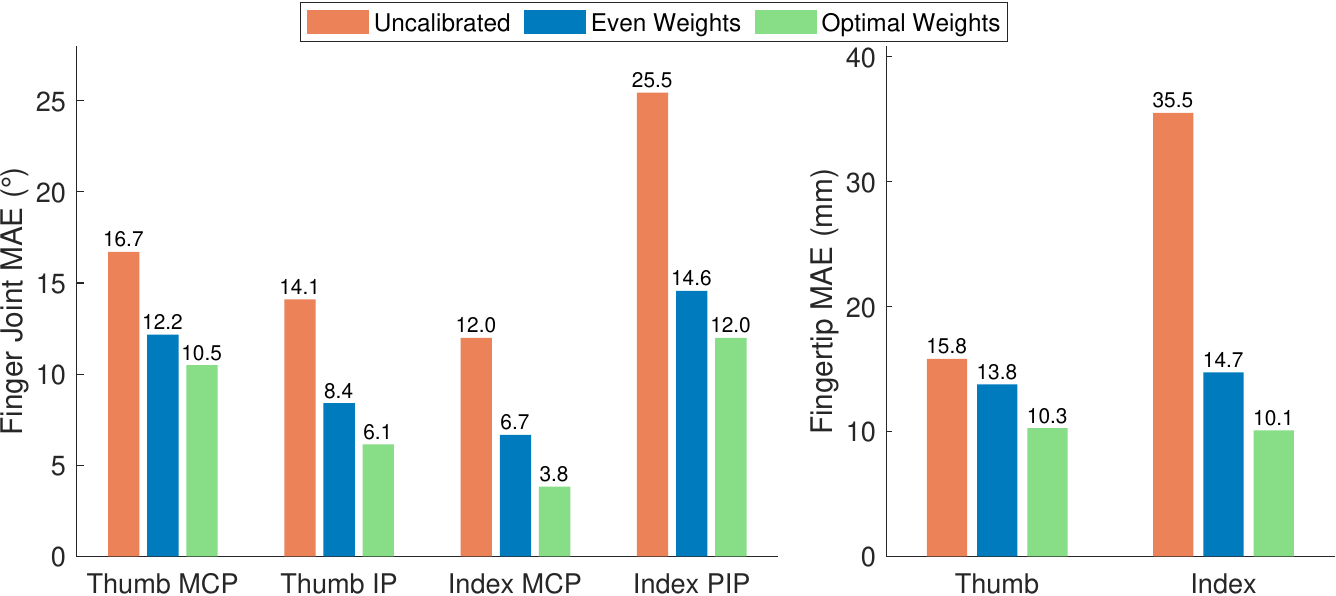}
    \caption{Hand tracking accuracy results averaged across subjects and tasks. (left) thumb and index joint angles errors  and (right) thumb and index fingertip position errors.}
    \label{fig:bar_graph_avg}
    \vspace{-1em}
\end{figure}
\begin{figure}[t!]
  \centering
  \begin{subfigure}[t]{0.95\linewidth}
    \includegraphics[width=\linewidth]{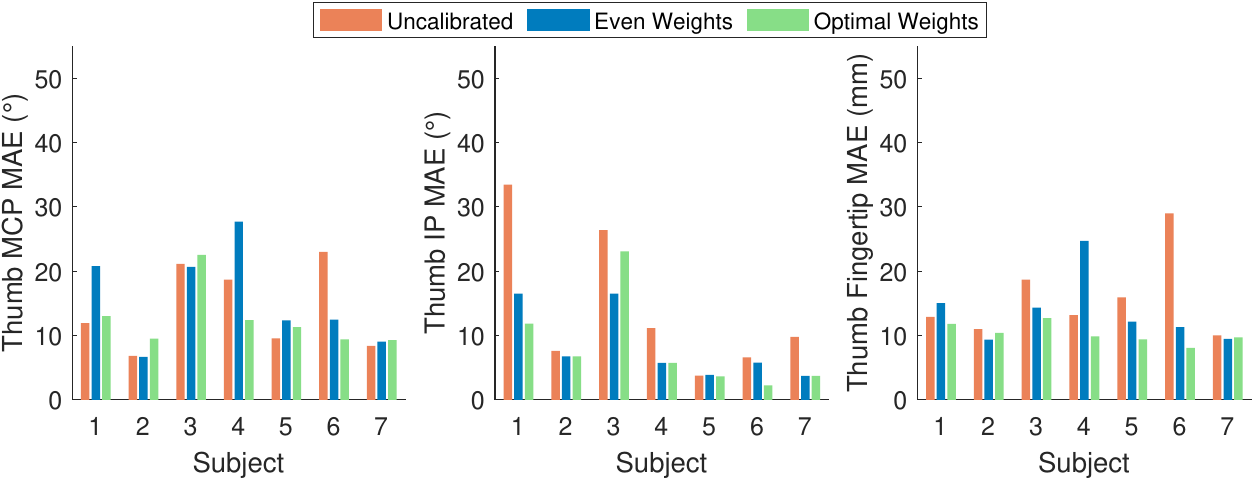}
    \caption{Combined Thumb MCP and IP Flexion}
        \label{subfig:avg}
    \end{subfigure}
        \par\medskip
        \begin{subfigure}[t]{0.95\linewidth}
          \includegraphics[width=\linewidth]{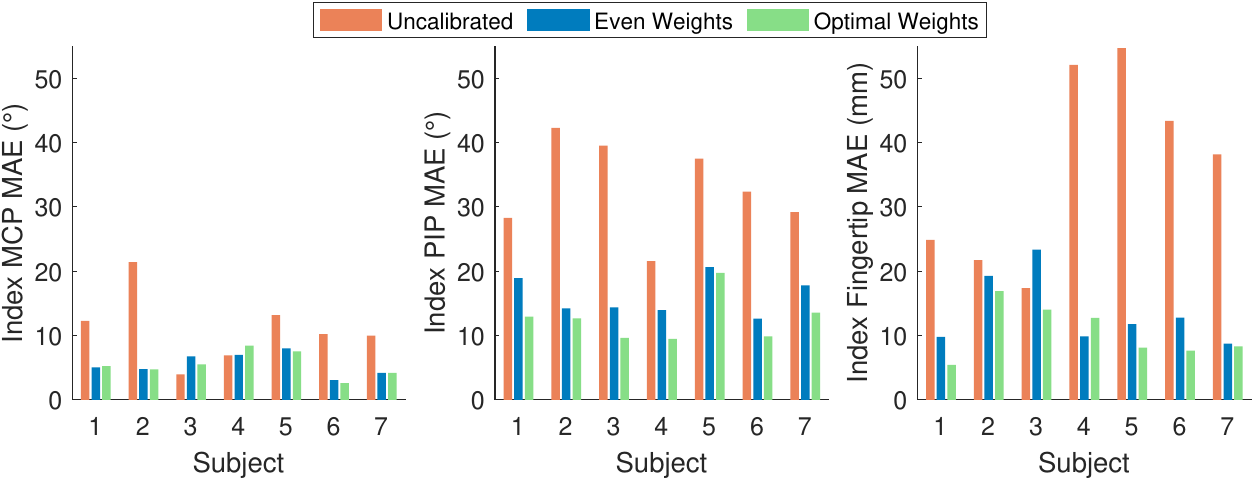}
    \caption{Combined Index MCP and IP Flexion}
                  \label{subfig:index1_bar}
    \end{subfigure}

  \caption{Hand-tracking accuracy results for 7 subjects. (a) Thumb MCP and IP joint angle errors and fingertip position errors (b) Index-finger MCP and PIP joint angle errors and fingertip position errors.}
  \label{fig:bar_graph_subjects}
  \vspace{-3mm}
\end{figure}

We report mean absolute errors (MAE) between the \textsc{Maestro}-based tracking and ground truth for joint angles and fingertip positions. Average results are shown in Fig.~\ref{fig:bar_graph_avg}, while subject-specific results for the three tracking tasks are shown in Fig.~\ref{fig:bar_graph_subjects}. 

\begin{table}[t]
\fontsize{6.5}{6}\selectfont
\setlength{\tabcolsep}{1.5pt}
\renewcommand{\arraystretch}{1.1}
\centering
\begin{tabular}{c|cc|cc|cc|cc|cc|cc|cc}
\toprule
\textbf{Sub}&\textbf{Length}&\textbf{Width}
& \multicolumn{6}{c|}{\textbf{Thumb (MAE \%)}} 
& \multicolumn{6}{c}{\textbf{Index (MAE \%)}} \\
&  & 
& \multicolumn{2}{c}{MCP} & \multicolumn{2}{c}{IP} & \multicolumn{2}{c|}{Tip} 
& \multicolumn{2}{c}{MCP} & \multicolumn{2}{c}{PIP} & \multicolumn{2}{c}{Tip} \\
& (cm) & (cm) 
& Even & Opt. & Even & Opt. & Even & Opt. 
& Even & Opt. & Even & Opt. & Even & Opt. \\
\midrule
1 & 17.0 & 6.5 
  & \textcolor{red}{-74.7} & \textcolor{red}{-9.3} 
  & 50.7 & 64.6 
  & \textcolor{red}{-16.3} & 8.7 
  & 59.1 & 57.2 
  & 33.2 & 54.3 
  & 60.7 & 78.1 \\
2 & 17.2 & 8.0 
  & 2.2 & \textcolor{red}{-25.5} 
  & 11.4 & 11.4 
  & 14.7 & 5.3 
  & 77.7 & 77.9 
  & 66.4 & 70.0 
  & 11.2 & 22.3 \\
3 & 18.0 & 9.5 
  & 2.3 & \textcolor{red}{-6.6}  
  & 37.5 & 12.5 
  & 23.3 & 32.1 
  & \textcolor{red}{-71.3} & \textcolor{red}{-39.0} 
  & 63.6 & 75.6 
  & \textcolor{red}{-34.4} & 19.3 \\
4 & 18.5 & 9.0 
  & \textcolor{red}{-48.3} & 43.6 
  & 48.6 & 48.6 
  & \textcolor{red}{-87.8} & 25.2 
  & \textcolor{red}{-0.8} & \textcolor{red}{-3.1} 
  & 35.4 & 56.1 
  & 81.1 & 75.6 \\
5 & 17.0 & 8.5 
  & \textcolor{red}{-29.4} & \textcolor{red}{-15.4} 
  & \textcolor{red}{-2.9} & 3.2 
  & 23.6 & 41.2 
  & 39.3 & 42.9 
  & 45.0 & 47.3 
  & 78.5 & 85.2 \\
6 & 19.2 & 9.8 
  & 45.8 & 69.1 
  & 12.7 & 66.1 
  & 61.1 & 72.1 
  & 70.0 & 74.5 
  & 61.0 & 69.5 
  & 70.6 & 82.4 \\
7 & 18.1 & 7.8 
  & \textcolor{red}{-7.9} & \textcolor{red}{-10.8} 
  & 62.1 & 62.1 
  & 5.3 & 3.0 
  & 58.2 & 58.2 
  & 39.0 & 53.6 
  & 77.1 & 78.3 \\
\midrule

\textbf{Avg} & \textbf{17.6} & \textbf{8.4} 
  & \textbf{26.9} & \textbf{37.1}
  & \textbf{40.4} & \textbf{56.7}
  & \textbf{44.2} & \textbf{68.3}
  & \textbf{42.7} & \textbf{52.9}
  & \textbf{12.7} & \textbf{34.8}
  & \textbf{58.6} & \textbf{71.5} \\
\bottomrule
\end{tabular}
\caption{Hand dimensions and calibration results for seven subjects. Entries show the percent change in joint/tip MAE under even-weighted (Even) and optimal-weighted (Opt.) calibration relative to the uncalibrated baseline.
}
\label{tab:results}
\vspace{-2.0em}
\end{table}

Table~\ref{tab:results} summarizes the results, showing joint-wise and fingertip position error reductions across all participants. Despite some variability across joints and subjects, calibration consistently led to notable reductions in both joint and fingertip tracking errors. The greatest improvements were observed in the thumb IP and index PIP joints, as well as in the fingertip position of the index finger. Fingertip position accuracy improved similarly, with the index fingertip achieving the highest average error reduction (71.5\%), while the thumb fingertip showed a more moderate improvement (34.8\%). These gains support the calibration’s effectiveness in enhancing end-effector representation, which is especially relevant for precision manipulation tasks.

The optimal-weighted calibration consistently yielded the lowest MAE, followed by the even-weighted calibration, while the uncalibrated model exhibited the highest errors across all tested joints. On average, the optimal-weighted approach produced greater error reductions, particularly at the fingertip level, with the index finger showing the most consistent improvements. The thumb also showed improvement following calibration, albeit with greater variability, which can be attributed to anatomical variability and the complexity of its joint articulation. Occasional negative error reductions in the thumb and index MCP joints may reflect residual alignment errors or unmodeled mechanical coupling. Overall, the consistent error reductions across participants suggest that the calibration improved tracking across participants with different hand sizes.


\begin{figure}[t!]
    \centering
    \vspace{0.5em}
    \begin{subfigure}[t]{0.95\linewidth}
        \includegraphics[width=\linewidth]{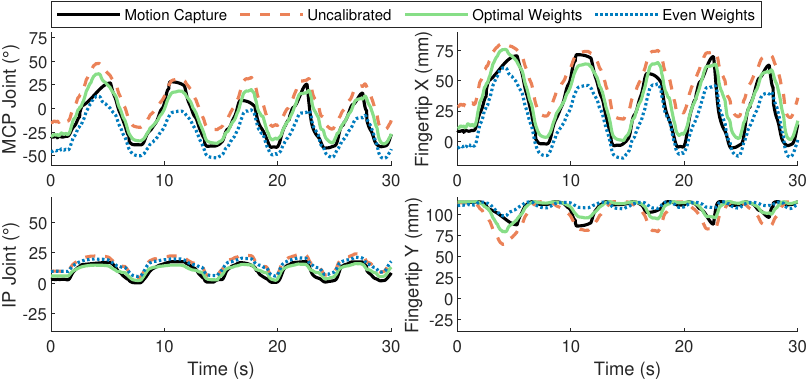}
        \caption{Combined Thumb MCP and IP Flexion}
    \end{subfigure}
    \par\medskip
    \begin{subfigure}[t]{0.95\linewidth}
        \includegraphics[width=\linewidth]{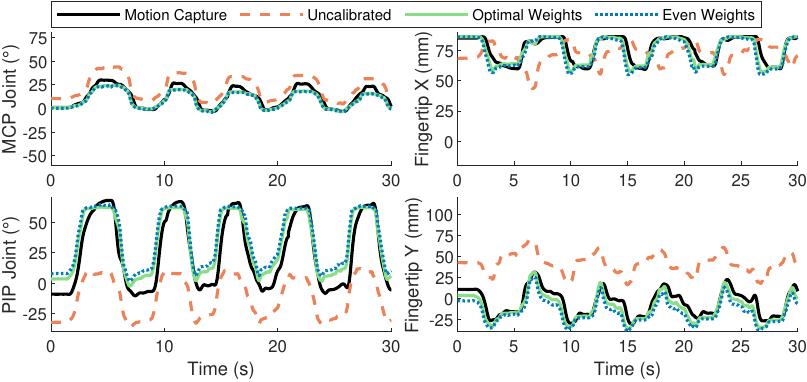}
        \caption{Combined Index MCP and PIP Flexion}
    \end{subfigure}
\caption{Comparison of hand-tracking performance for \textit{Subject 6}. Motion-capture trajectories are shown alongside uncalibrated and calibrated estimates using even and optimal weights.}    \label{fig_9:results}
    \vspace{-0.5em}
\end{figure}

As shown in Fig.~\ref{fig_9:results}, for \textit{Subject 6}, the weighted calibration achieves trajectories that align most closely with the ground truth, outperforming both the uncalibrated and even-weighted models. This example highlights the method’s effectiveness in reducing tracking errors during dynamic finger motion.

\subsection{Qualitative Results}
\begin{figure}[t!]
  \centering
   \vspace{-1em}
  \begin{subfigure}{0.42\columnwidth}
    \includegraphics[width=\linewidth]{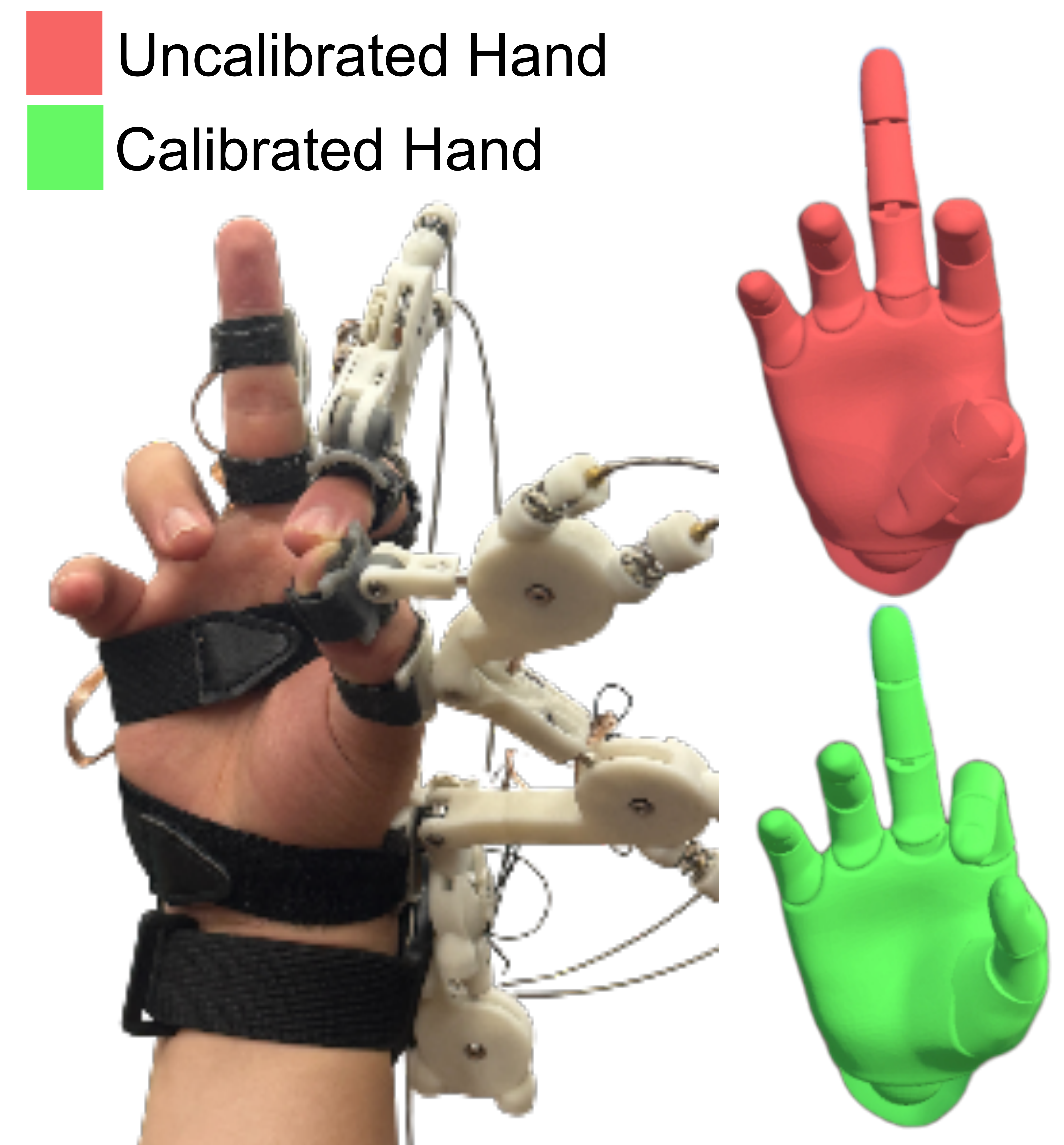}
    \caption{Index finger pinch}
    \label{subfig:index_pinch}
  \end{subfigure}%
  \begin{subfigure}{0.42\columnwidth}
    \includegraphics[width=\linewidth]{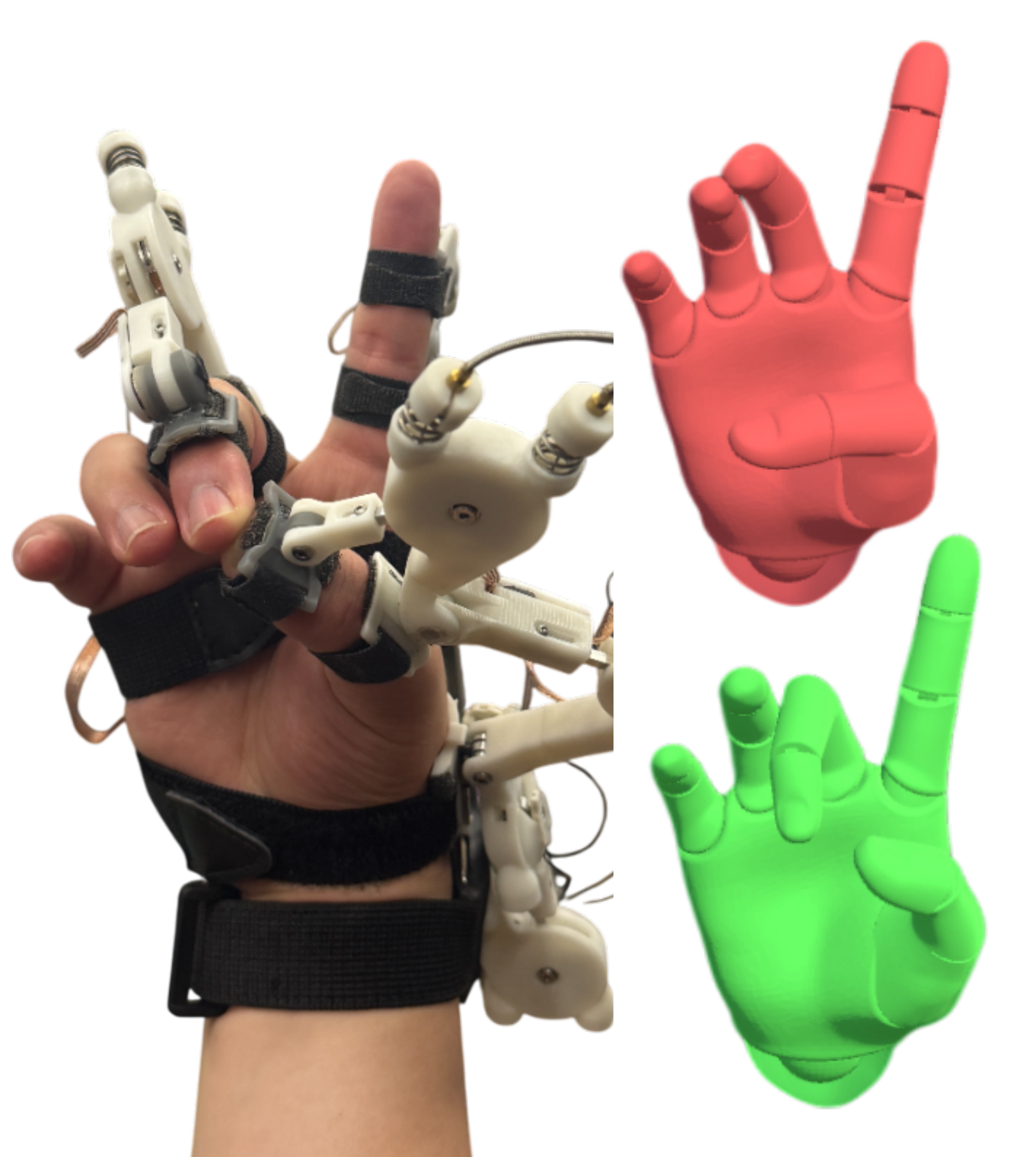}
    \caption{Middle finger pinch}
    \label{subfig:middle_finger_pinch}
  \end{subfigure}

  \caption{Real and virtual hands (Unity rendering) shown side by side for uncalibrated (red) and calibrated (green) tracking.}
  \vspace{-1em}
  \label{fig_10:visual}
\end{figure}

To qualitatively assess tracking improvements, we visualized results using a virtual Unity hand, driven by joint estimates under calibrated and uncalibrated conditions. Representative postures (e.g., pinch) were recorded for each subject to visually evaluate tracking fidelity. As shown in Fig.~\ref{fig_10:visual}, red hands represent uncalibrated data, while green hands show calibrated poses. Calibration visibly improves alignment with the real hand, especially in fine thumb and finger movements, qualitatively confirming the enhanced motion fidelity. A demonstration of real-time virtual hand tracking is provided in the \textit{supplementary video}.

%% file: 05_discussion.tex
\section{Discussion}
\label{sec:discussion}

This study introduced a subject-specific calibration framework to improve hand-tracking accuracy in a sensorized robotic exoskeleton. As shown in Section~\ref{sec:results}, calibration reduced both joint angle and fingertip errors across participants with varying hand sizes and donning conditions, confirming the method’s effectiveness. An optimal weight distribution further improved performance, particularly in joints influenced by closed-loop kinematics, confirming that user-specific variability and misalignment must be explicitly calibrated by prioritizing kinematic parameters that are more sensitive to different hand sizes or slippage. 

Previous studies have also examined calibration trade-offs and shape-specific adaptation in data-glove systems~\cite{heinrich2024comparison,li2025fsglove}, which align with our findings and further emphasize the need for efficient, subject-agnostic calibration strategies.

Our framework applies this concept to robotic exoskeletons by embedding subject-specific calibration directly into the device’s kinematic model. By optimizing virtual parameters from redundant sensing and reference hand poses, it achieves subject-specific kinematic alignment without additional instrumentation or repetitive manual calibration.

In terms of performance, the proposed method achieves a mean absolute fingertip error of approximately 10 mm, comparable to or better than markerless vision-based systems that often exceed 10–20 mm~\cite{abdlkarim2024methodological}. Prior studies have reported tolerable fingertip errors up to 10-15 mm without perceptible loss of control fidelity~\cite{handa2020dexpilot,leonardis2024hand}, supporting the feasibility of the proposed framework for teleoperation tasks.

The data-driven weight search revealed similar trends across participants, suggesting that an averaged profile is a reasonable initialization, though device- and user-specific refinement remains necessary for accurate tracking. 

While demonstrated on the \textsc{Maestro} hand exoskeleton, the calibration framework can be generalized to other wearable systems that provide comparable independent kinematic constraints. While in this study geometric loop closures enabled solving joints without redundant sensing (e.g., the thumb IP), the optimization remains well-posed as long as the number of identifiable parameters does not exceed the available independent constraints.

Despite the promising results, the calibrated model does not remove all discrepancies: residual joint-angle and fingertip errors persist due to model mismatch, sensing noise, and unmodeled compliance, and are most noticeable at the fingertips, where small joint deviations accumulate along the kinematic chain. A complementary fingertip-based or task-level calibration can therefore be used to further refine fingertip alignment for dexterous tasks that demand higher accuracy.


This study has several other limitations. First, although the evaluation included continuous finger motion, the analysis primarily focused on short, isolated movement segments rather than long-duration or coordinated multi-finger actions, limiting insight into more complex manipulation behaviors. Second, we acknowledge that the participants represent hand sizes slightly below the population mean, highlighting the need for validation across a broader and more diverse range of users. Third, the evaluation assessed relative error reduction rather than benchmarking against other devices or calibration methods; establishing unified accuracy standards will be an important direction for future work.

Future work includes online or self-adaptive calibration to account for changes in fit or sensor drift, as well as extensions to simultaneous multi-finger calibration to further improve dexterity modeling and real-world usability across diverse robotic hands.


\section{Conclusion}
In conclusion, this work presents a subject-specific calibration framework that improves the accuracy of exoskeleton-based hand tracking for teleoperation. Implemented on the \textsc{Maestro} hand exoskeleton, the calibration significantly reduces tracking errors by optimizing virtual link parameters and residual weights. While residual joint-angle and fingertip discrepancies remain, the method improves motion fidelity and can support teleoperation tasks.
By optimizing virtual link parameters and residual weights, it achieves consistently higher hand tracking fidelity. Although residual joint-angle errors of about 10\textdegree{} remain, the calibration enables precise tracking for fine manipulation and teleoperation tasks. Rather than being universally applicable, the approach provides a structured calibration procedure that can be adapted to other hand–exoskeleton designs with similar kinematic constraints.